\documentclass[10pt,journal,compsoc]{IEEEtran}
\ifCLASSOPTIONcompsoc
  \usepackage[nocompress]{cite}
\else
  \usepackage{cite}
\fi

\ifCLASSINFOpdf
  \usepackage[pdftex]{graphicx}
\else
\fi

\usepackage{amsmath}
\usepackage{amsfonts}
\usepackage[svgnames,table]{xcolor} 
\usepackage{multirow}
\usepackage{float}
\usepackage{url}
\usepackage{footnote}

\newcommand{\etal}{\textit{et al}.}

\begin{document}
%
\title{Real-time Keypoints Detection for Autonomous Recovery of the Unmanned Ground Vehicle}
%
%
%
%

\author{Jie~Li$^{1}$, 
        Sheng~Zhang$^1$, 
        Kai~Han$^2$, 
        Xia~Yuan$^{1}$, 
        Chunxia~Zhao$^1$, 
        and~Yu~Liu
\IEEEcompsocitemizethanks{
\IEEEcompsocthanksitem X. Yuan is the corresponding author.
\IEEEcompsocthanksitem J. Li, S. Zhang, X. Yuan and C. Zhao are with Nanjing University of Science and Technology, Nanjing, China.
E-mail: yuanxia@njust.edu.cn
\IEEEcompsocthanksitem K. Han is with University of Oxford, Oxford, United Kingdom. E-mail: kai.han@bristol.ac.uk
\IEEEcompsocthanksitem Y. Liu is with The University of Adelaide, Adelaide, SA, Australia.
}
}

%
%

\markboth{
Journal of \LaTeX\ Class Files,~Vol.~14, No.~8, August~2015
}%
{Shell \MakeLowercase{\textit{et al.}}: Bare Demo of IEEEtran.cls for Computer Society Journals}

\IEEEtitleabstractindextext{%
\begin{abstract}
The combination of a small unmanned ground vehicle (UGV) and a large unmanned carrier vehicle allows more flexibility in real applications such as rescue in dangerous scenarios.
The autonomous recovery system, which is used to guide the small UGV back to the carrier vehicle, is an essential component to achieve a seamless combination of the two vehicles.
This paper proposes a novel autonomous recovery framework with a low-cost monocular vision system to provide accurate positioning and attitude estimation of the UGV during navigation. 
First, we introduce a light-weight convolutional neural network called UGV-KPNet to detect the keypoints of the small UGV form the images captured by a monocular camera. UGV-KPNet is computationally efficient with a small number of parameters and provides pixel-level accurate keypoints detection results in real-time. 
Then, six degrees of freedom pose is estimated using the detected keypoints to obtain positioning and attitude information of the UGV.
Besides, we are the first to create a large-scale real-world keypoints dataset of the UGV.
The experimental results demonstrate that the proposed system achieves state-of-the-art performance in terms of both accuracy and speed on UGV keypoint detection, and can further boost the 6-DoF pose estimation for the UGV.

\end{abstract}

}

\maketitle

\IEEEdisplaynontitleabstractindextext

%
\IEEEpeerreviewmaketitle


\section{Introduction}
With the rapid development of robotics in recent years, the unmanned ground vehicles (UGV) empowered with the ability of autonomous navigation has received increasing attention
in the field of robotics~\cite{luo2018dynamic,veloso2015cobots,omrane2016fuzzy,sun20183dof}.
On the one hand, there are already many exciting achievements at present. 
For example, self-driving cars~\cite{paden2016survey} are making progress to levels four and five of driver assistance technology advancements and are merging onto roadways.
The sweeping robot with autonomous navigation and recharge function~\cite{kurazume2000development,luo2008bioinspired,hasan2014path,yakoubi2016path} is helping millions of households reduce the burden of daily cleaning.
Some inspection robots working in a relatively closed environment can automatically return to the base according to the established route as well~\cite{guarnieri2009helios,baranzadeh2017distributed}.
On the other hand, many situations require the cooperation of large and small vehicles, such as long-distance field rescue, environmental inspection, and hazardous site surveys.
Although the small UGV has a short operating range due to the limitation of speed and battery life, it has the advantages of flexibility and compactness.
Large vehicles have high speed and long-range, but they cannot be used in complex situations with limited space. 
Therefore, the combination of small UGVs and large carrier vehicles is a more plausible choice by combining the benefits of both.
For example, the combination of a small and a large vehicles can effectively rescue people and animals trapped due to an earthquake or fire. 
The small UGV can be quickly transported to the vicinity of the disaster by the large carrier vehicle, and take advantage of its small size to go deep into the disaster site for rescue.
Then, the recovery system guides the UGV to board the carrier vehicle automatically. After that, the carrier vehicle works for quick access to treatment for the injured.
The automatic deployment and recovery process is a vital component of combining the small UGV and the carrier vehicle. 
The process is carried out in a place, where there is enough space for both the small UGV and the carrier vehicle to perform the deployment, near the mission site.
When the UGV has completed its task and is ready to return to the carrier vehicle, accurate navigation is required to ensure a successful recovery. In some applications, both the UGV and the carrier vehicle are assumed to be self-driving, which requires real-time automated navigation.
Several methods have yielded good results~\cite{luo2018dynamic,veloso2015cobots} for specific closed environments, limiting their wide application. 
In contrast, the proposed system is expected
to work well in challenging open situations with both indoor and outdoor scenes.
The open environment in the wild has more uncertain factors and requires UGV to have stronger environmental recognition and navigation capabilities~\cite{pandey2017mobile} compared with the relatively closed environment. 
\cite{li2017automatic} introduces an approach that sets up a lidar on the small UGV to do the location by detecting the target.
Due to the limitation of the field of view, the target often exceeds the field of view of the lidar, resulting in unstable detection results.
Besides, lidars are expensive for small UGVs.
A vision system using a camera is a low-cost solution and can provide positioning information by detecting the bounding box of the target.
However, the bounding box cannot provide high-precision positioning information and cannot address the problem of attitude estimation.
Also, the real-time requirement of image-based vision systems is a big challenge.


%
%
\begin{figure}[!tbp]
\centering
{
\includegraphics[width=0.95\linewidth]{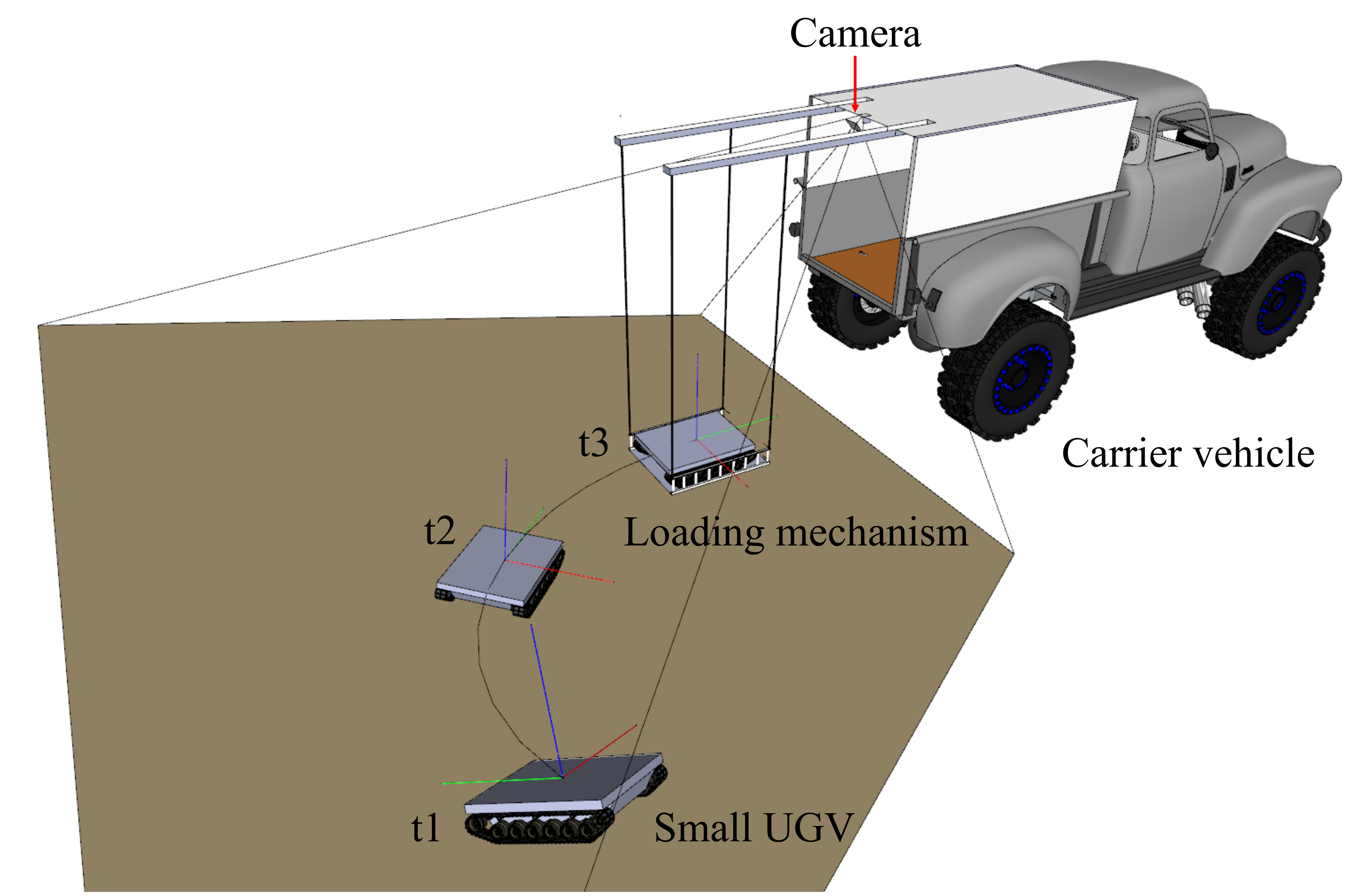}
}
\caption{
\textbf{Autonomous Recovery System.}
The system uses a camera mounted on the top of the carrier vehicle to capture the images for the UGV. Our UGVPoseNet takes these images as input to detect keypoints of the UGV. With the detected keypoints, the 6-DoF pose can then be estimated by solving the PnP problem. Next, the system can guide the small UGV to move from one time step to another (e.g., $t_1$ to $t_2$, and $t_2$ to $t_3$). 
Finally, the loading mechanism lifts the small UGV and pulls it into the carrier vehicle automatically.
}
\label{fig:teaser}
\end{figure}

In this paper, we focus on the task of UGV recovery and provide a novel and practical approach based on a low-cost monocular vision system.
As shown in Fig.~\ref{fig:teaser}, the entire system consists of a ground carrier vehicle, a loading mechanism, and a small unmanned ground vehicle (UGV). During the autonomous recovery process to pick up the UGV, a camera mounted on the carrier vehicle continuously provides images of the UGV within the field of view.
The captured images are processed by the proposed convolutional neural network called UGV-KPNet to perform positioning and attitude estimation.
The pose information is then sent to the control module to navigate the UGV to return to the hanging frame automatically.
Finally, the loading mechanism retracts the hanging frame and load the small UGV into the carrier vehicle.
The whole system mainly relies on the images captured by the monocular camera to obtain information. 
The system is practical and easy to deploy and is more cost-effective than the laser and lidar solution.
The carrier vehicle provides a long time cruising ability and can extend the range of the UGV operating as a relay. The UGV has more nimble mobility and a smaller profile, which helps it to perform tasks in hazardous areas.
The large carrier vehicle and the small UGV work like a fire engine and firefighters. The fire engine transports firefighters to the scene but will not drive directly into it. 
Compared with the working area of the UGV, the surrounding environment of the carrier vehicle is relatively safe, open, and accessible. In the paper, the proposed method is designed for the stage after the UGV completing specific tasks. In particular, the autonomous recovery system guides the small UGV back to the carrier vehicle and loads the small UGV for transportation. Thus, the autonomous recovery process is conducted in relatively open areas.
The six degrees of freedom (6-DoF) pose (three for translation and three for rotation) estimation of the UGV is one of the keys to the autonomous recovery system, which provides precise localization and attitude information of the UGV.
In the system, the 6-DoF estimation is based on the keypoints detection of the UGV and solving the Perspective-n-Point (PnP) problem.
First, we propose UGV-KPNet for the keypoints detection of the UGV, which is a light-weight convolutional neural network and can provide accurate detections in real-time.
In UGV-KPNet, 
a series of effective improvements are made to obtain high-performance keypoint prediction results.
Compared with previous methods, 
UGV-KPNet achieves high detection accuracy with a small number of parameters and low computational costs.
Meanwhile, it can quickly predict reliable keypoints to ensure UGV's high-precision positioning and real-time navigation.
Then, we address the PnP problem by using Levenberg-Marquardt optimization~\cite{roweis1996levenberg} to get the 3D position and attitude of the UGV.
The detected keypoints of the UGV and their relationships are provided as a priori information to solve the PnP problem.
The height of the UGV and the height and attitude of the camera are also fixed and known.
By adopting the processes sequentially, we can obtain the accurate 6-DoF pose of the UGV in real-time from the monocular vision.

Due to the absence of training data, 
we create UGVKP, a large-scale UGV keypoint dataset with various scenes in both indoor and outdoor environments.
Each image in UGVKP is manually annotated with four keypoints of the UGV.
To the best of our knowledge, this is the first keypoint dataset for UGV.
The dataset and code are available at https://waterljwant.github.io/UGV-KPNet/.

The main contributions of this paper can be summarized as follows:
\begin{itemize}
    \item 
    This paper proposes a novel autonomous recovery framework consisting of a small UGV, a loading mechanism, and a ground carrier vehicle with a low-cost monocular vision system.
    \item 
    An effective light-weight convolutional neural network called UGV-KPNet for UGV keypoints detection is proposed. It has high detection accuracy with a small number of parameters and a low computational cost while predicting reliable keypoints in real-time.
    \item 
    This paper introduces a practical and low-cost 6-DOF pose estimation solution that includes keypoint detection and PnP problem solving.
    \item 
    The UGVKP dataset is created, which is a real-world, large-scale dataset of UGV containing both challenging indoor and outdoor scenes with keypoints annotations, to provide the training data for the neural network.
\end{itemize}

\section{Related Work}
\subsection{Ground Vehicle Navigation}

At present, ground vehicle navigation mainly relies on the Global Positioning System (GPS)~\cite{abbott1999land,stanvcic2010integration}.
However, GPS receivers often suffer from the lack of positioning accuracy, availability, and continuity due to the insufficient number of visible satellites in the tunnel, indoor environments, and multipath errors in the urban area~\cite{lim2017integration,chiang2013performance}.
Current civilian GPS positioning technology is difficult to achieve centimeter-level positioning accuracy~\cite{yuan2017estimation}, and therefore cannot meet the high-precision requirements of small UGV positioning.
Therefore, GPS is suitable for navigation in a large area but not for the precise guidance of the small ground vehicle during the autonomous recovery process.

Lidar can provide high-precision ranging information, which relies on structured information as the feature to sense the environment.
Lidar has a wide range of applications in the field of autonomous driving, especially map creation and positioning.
Li~\etal~\cite{li2017automatic} adopt a single-line lidar (2D lidar) mounted on the UGV horizontally, and use it to detect a preset landmark on the carrier vehicle for positioning.
This method can be used during the day and night to achieve high accuracy positioning.
However, affected by terrain fluctuations, it is difficult for the single-line lidar to capture the target landmark.
Besides, the valid detection distance of the lidar is short, since the active points are sparse due to divergence.
Also, it is expensive to equip lidar for small UGVs.

Compared to lidar, using the camera is cost-saving and can provide a better field of view.
In this paper, we propose a vision-based method to provide information to locate the small UGV accurately.
We set a camera on the top of the carrier vehicle to obtain a large field of view and use the images acquired by the camera to estimate the pose of the UGV.
To obtain high-precision detection results, we propose a deep learning method that uses a convolutional neural network (CNN) to extract image features and get the positions of keypoints.
To achieve real-time detection, we designed a lightweight neural network, which has a smaller amount of parameters and a lower computational cost than the previous method.
To ensure stable and accurate detection results, we have carefully designed multiple novel modules of the network structure.
The camera is deployed on the carrier vehicle, reducing not only the load but also the computational cost on the UGV. These improvements save UGV power consumption and extend its battery life.

\begin{figure*}[ht]
\centering
{
\includegraphics[width=\linewidth]{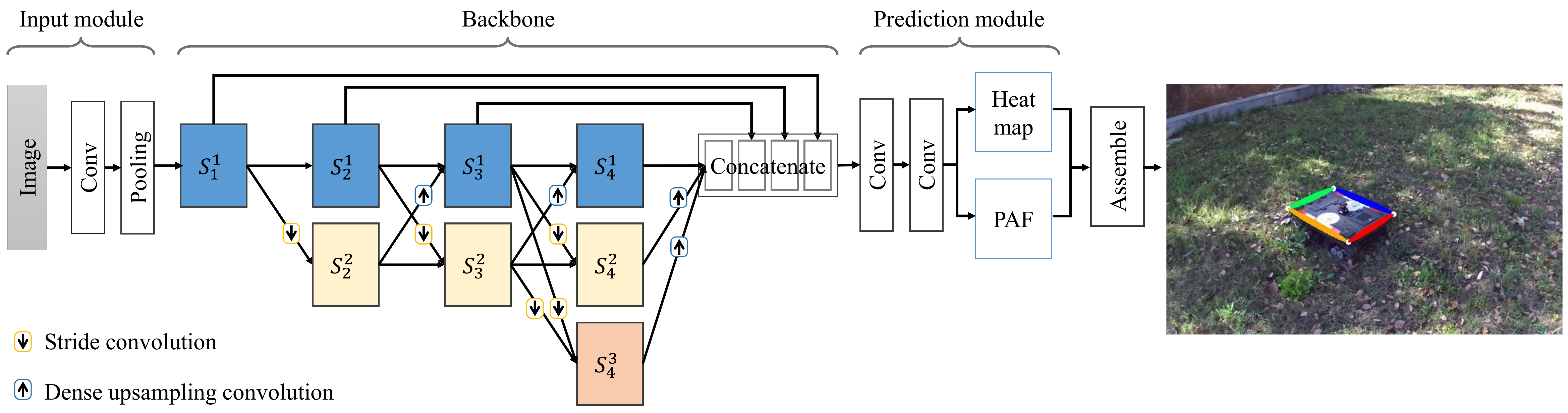}
}
\caption{
\textbf{UGVPoseNet: keypoints detection network for UGV pose estimation.}
Taking a image captured by the monocular camera mounted on the top of the carrier vehicle as input, the network predicts the heatmaps of the four keypoints and the PAF of the connected points of the UGV. 
The assemble module combines the candidate keypoints in the heatmap and the PAFs to obtain ordered structure of the UGV.
}
\label{fig:NetworkStructure}
\end{figure*}

\subsection{Keypoints Detection via CNNs}
In recent years, with the development of deep learning technologies~\cite{krizhevsky2012imagenet,Simonyan15}
and the emergence of large-scale data sets~\cite{ILSVRC15,lin2014microsoft}
in target detection~\cite{he2017mask,redmon2017yolo9000,tian2019fcos},
keypoint detection~\cite{papandreou2017towards,OpenPoseTPAMI,sun2019deep}
environmental understanding~\cite{Li2019ddr,palnet}
and other fields have achieved remarkable results.
The keypoint detection technology that this paper focuses on has a prominent effect in the field of face recognition.
The approach in~\cite{wu2017facial} can provide high-quality facial landmark detection.
However, this method cannot detect the region of interest and relies on additional operations to provide detected faces as input.
%
\cite{papandreou2017towards} proposes a two-stage keypoints detection method.
First, the person in images is detected by predicting the bounding box. Then the keypoint location is performed within the bounding box.
\cite{wu2018simultaneous} uses a two-stage strategy to estimate the 3-DoF pose of a human face as well.
The first stage predicts the bounding box of the detected face.
Then, 3D pose estimation can be performed directly on the face within the bounding box without using landmarks.
However, this method can only estimate the 3-DoF pose of a human face.
The bounding box can not provide accurate 3D position information.

In the field of object keypoints detection, \cite{pavlakos20176} estimate the continuous six degrees of freedom (6-DoF) pose (3D translation and rotation) of an object from a single image.
However, this method is still a two-stage method, which uses Faster RCNN~\cite{ren2015faster} to detect the bounding box of objects, and then performs keypoints localization and pose optimization.
Besides, the 6-DoF estimation relies on the corresponding 3D models of the objects.

There are also many useful methods for the detection of keypoints in the human body. 
\cite{kocabas2018multiposenet} proposes a network with two branches to perform keypoint heatmap prediction and human bounding box detection, respectively.
However, this complicated network cannot work in real-time.
One-stage methods such as OpenPose~\cite{OpenPoseTPAMI} do not need the extra operation to detect the bounding box of the targets. Instead, OpenPose directly locates keypoints and use the part association filed to assemble the keypoints and the joints of each person.
However, this method has a large number of parameters and high computational costs.
It relies on powerful computing resources to make predictions in real-time.
However, computing resources of small UGVs are limited.

To address the computational limitation,~\cite{guan2019realtime} propose the lightweight OpenPose model by replacing the backbone from VGG19~\cite{Simonyan15} to shufflenetv2~\cite{he2016deep}.
This method reduces the number of network parameters and calculations significantly.
However, this method also weakens the ability of the network's feature representation, leading to a decrease in detection accuracy.

In~\cite{chen2018cascaded}, cascaded pyramid network is proposed,
which takes the advantages of the multi-level features to improve the prediction accuracy of difficult points.
However, this method also suffers from a large number of parameters and calculations.

This paper proposes the UGV-KPNet for the keypoint detection of the UGV. 
The proposed UGV-KPNet has a lightweight structure with a small number of parameters and low computational costs as well.
Moreover, this network guarantees detection accuracy through a well-designed structure.
With the predicted keypoints, the 6-DoF pose of the UGV can be estimated. 

After obtaining the keypoints of the UGV, we estimate the UGV's 6-DoF in 3D space by solving the 3D-2D Perspective-n-Point (PnP) problem, where $n$ is the number of keypoints of the UGV in the system.
PnP has been a well-studied problem with excellent solutions such as Levenberg-Marquardt optimization~\cite{roweis1996levenberg} and EPnP~\cite{lepetit2009epnp}.

\section{Methodology}

The proposed autonomous recovery system is shown in Fig.~\ref{fig:teaser}, which consists of a small UGV, a carrier vehicle, and a lifting mechanism. 
In the system, the small UGV is a self-designed crawler ground mobile vehicle.
A camera is mounted at the rear of the carrier vehicle to capture images of the UGV.
We define the four vertices on the UGV shell as four keypoints. 
Then we use the proposed UGV-KPNet~(Section \ref{sec:keypoints_detection}), which is a lightweight convolutional neural network,
to detect the keypoints when the small UGV is in the field of view of the camera.
The four keypoints are used to solve the PnP problem by adopting the Levenberg-Marquardt optimization~\cite{roweis1996levenberg} to obtain the 6-DoF pose~(Section \ref{sec:6dof}) of the UGV.
With the real-time 6-DoF pose information, the UGV is navigated to the hanging board of the lifting mechanism. 
Finally, the lifting mechanism loads the UGV into the carrier vehicle, completing the entire autonomous recovery process.

The position and attitude estimation of the UGV is the core algorithm of the recovery system.
This paper proposes a vision-based approach to get the high accurate 6-DoF pose of the UGV. 
Our approach contains two phases, the detection of the UGV's keypoints and the 6-DoF pose estimation by solving the PnP problem to get the 3D position and attitude of the UGV.
In the first phase, we adopt the deep learning technology and design a convolutional neural network UGV-KPNet to detect the keypoints of the UGV.
We introduce improved shuffle block and dense upsampling convolution to build several multi-resolution subnetworks to form the backbone network for UGV-KPNet.
These innovations make UGV-KPNet a lightweight network and can detect the keypoints of UGV in real-time. Compared with previous methods, it requires significantly less computing resources. 
UGV-KPNet can also provide high-quality detection results with both high precision and recall scores.
The keypoints detected in this phase correspond to the four corners of the UGV's upper surface.
In the second phase, we obtain the 6-DoF pose estimation results by solving the 3D-2D PnP problem.
Since the 3D size of the UGV is known, the physical position relationship of the four keypoints on the UGV can be obtained based on this information.
We use these four pairs of 2D-3D correspondences to solve the positional relationship between the UGV and the camera.

During the recovery process, more accurate pose estimation is required when the UGV is close to the carrier compared to the case when the UGV is further away from the carrier. 
This is inherently true due to the nature of perspective projection used in the vision system.

\section{UGV Keypoints Detection}\label{sec:keypoints_detection}
This paper proposes UGV-KPNet which is a light-weight convolutional neural network to predict the keypoints of the UGV.
UGV-KPNet is deployed on the carrier vehicle and provide real-time prediction with a monocular camera to capture images. 
The proposed autonomous recovery method based on keypoint detection is performed in a scenario where both small UGV and carrier vehicle can access, and has sufficient field of view to capture the small UGV on site.
With a single image of UGV as input, UGV-KPNet predicts the heat map of the keypoints and the Part Affinity Field (PAF) of the connected points of the UGV. 
The heat map is a confidence map of keypoints to provide positions of the candidate keypoints.
Each candidate keypoint and the PAFs are then assembled to form an ordered structure to represent the UGV.
In the following sections, we introduce the details of this network.

\subsection{Network Structure}
The network structure of the proposed UGV-KPNet is shown in Fig.~\ref{fig:NetworkStructure}.
The captured image is first fed into the input module with a convolution layer and a pooling layer.
The main body of the network consists of four stages which are built with subnetworks of different resolutions~(Section \ref{sec:subnet}).
Several improved shuffle blocks~(Section \ref{sec:shuffleblock}) are stacked in the subnetworks.
Adjacent subnetworks of different resolutions are connected by stride convolution and dense upsampling convolution~(Section \ref{sec:DUC}).
The multi-level feature maps from different stages are concatenated and fed into the followed prediction module to predict the heat map~(Section \ref{sec:Heatmap}) and the Part Affinity Field~(Section \ref{sec:PAF}) through the residual adapter~(Section \ref{sec:adapter}).
The assemble module~(Section \ref{sec:assemble}) uses the relationships among the keypoints 
and the PAFs of the connections to output the ordered structure of the UGV.

%
%
\begin{figure}[!tbp]
\centering
{
\includegraphics[width=0.90\linewidth]{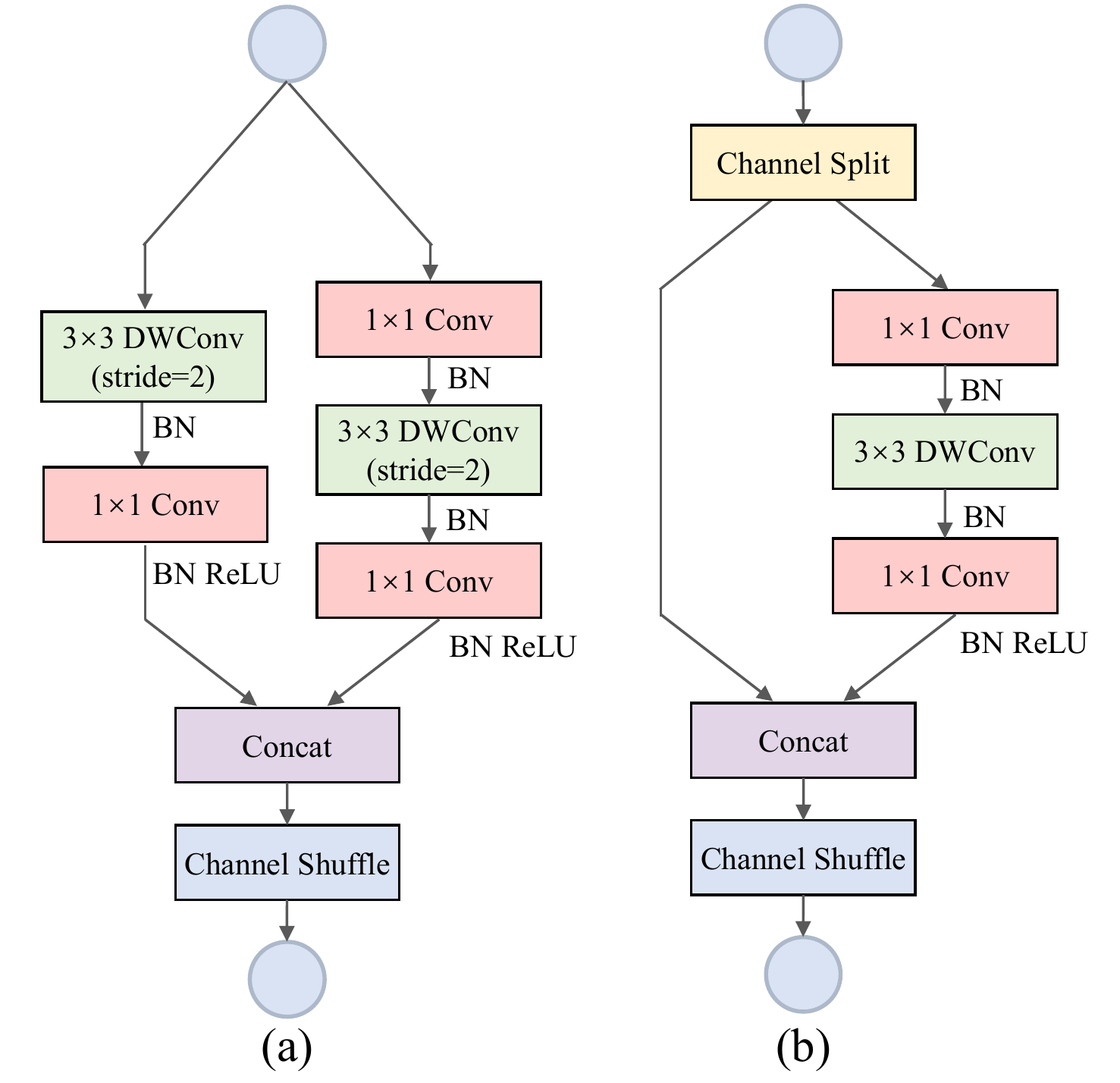}
}
\caption{
\textbf{Improved shuffle block.}
In each stage within the backbone, (a) is used as the first block and followed by several (b) basic blocks.
}
\label{fig:block}
\end{figure}


\subsubsection{Multi-resolution subnetworks}\label{sec:subnet}
The four stages of the backbone network are built with multi-resolution subnetworks.
The first stage has one subnetwork; each of the other stages consists of two or three subnetworks that are in different resolutions.
We denote the four stages as ${ S }_{ i },(i=1,2,\cdots ,4)$, and the subnetworks in each stage as $S_i^{j},(j=1,2,3)$.
The first subnetworks $S_i^{1}$ at the top of each stage keeps the same resolution as that of the previous feature map.
The feature map of the lower subnetwork in each stage is half the width and height of the upper subnetwork.
The output of these subnetworks with different resolutions are transferred through a down-sampling layer to perform the high-to-low connection and an upsampling layer to achieve the low-to-high connection.
As a result, each subnetwork in a later stage receives the information form all the subnetworks in the previous stage.
The multi-resolution representation is conducive to predicting the keypoint heat maps with accurate categories and precise spatial locations.

\subsubsection{Multi-level connection}\label{sec:multi-level_connection}
The skip-connections (solid lines above the four stages) from the stages with the same resolution subnetworks combines the low-level and high-level features, which are served as the input for the subsequent convolution layers.
The combined features from the multi-level connection can provide sufficient high-level semantic information to identify the category of the keypoints, and offer accurate low-level positioning information to determine the location of keypoints as well.

%
%
\begin{figure}[!tb]
\centering
{
\includegraphics[width=0.8\linewidth]{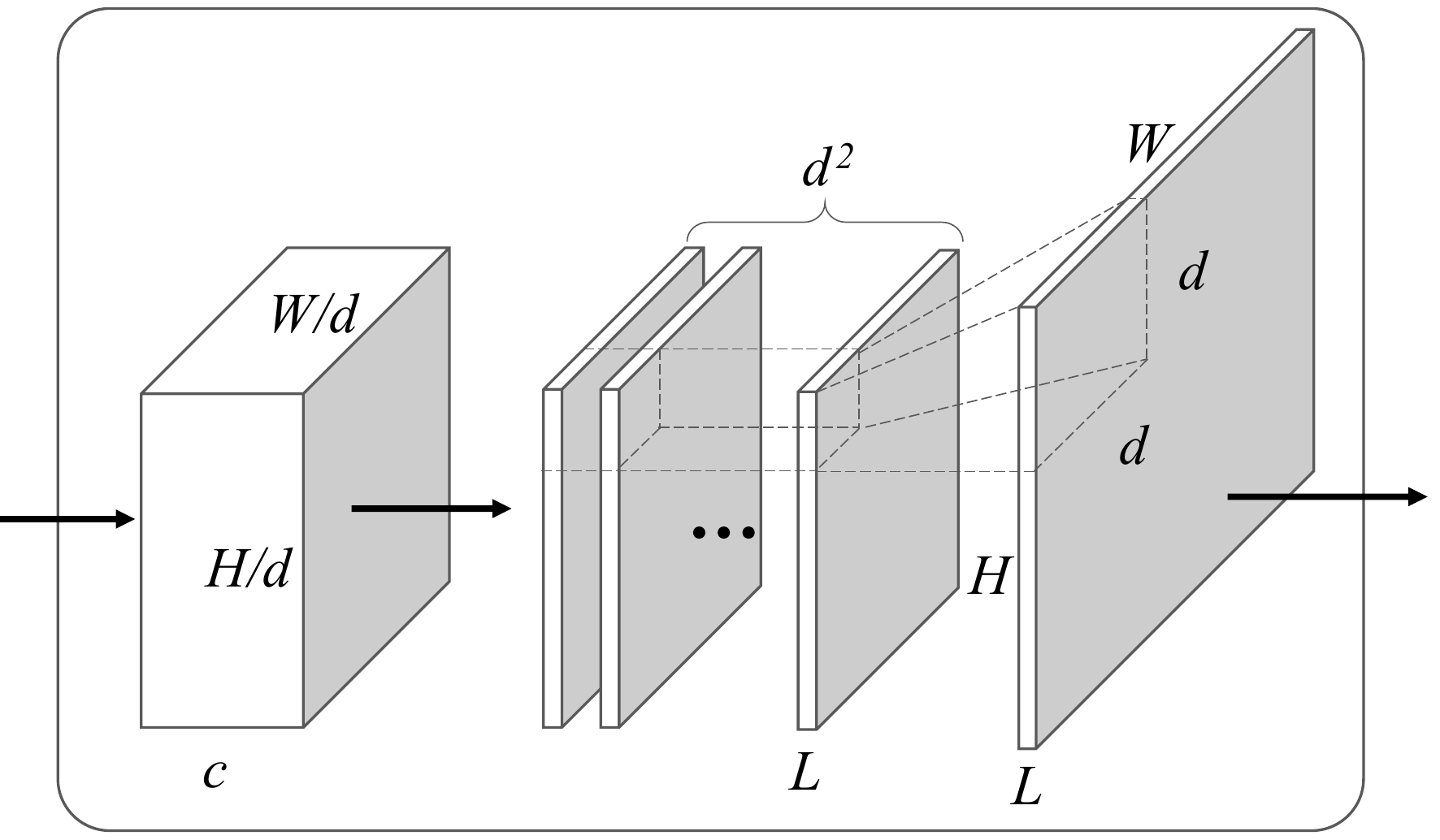}
}
\caption{
\textbf{Dense Upsampling Convolution (DUC).} 
The DUC operation is performed on the feature map of dimension $h \times w \times c$ to get the output feature map of dimension $h \times w \times (d^2 \times L)$, where $L$ is the number of channels in the high-resolution feature map. 
The output feature map is then reshaped to $H \times W \times L$.
where $h = H/d$, $w = W/d$, and $d$ is the upsampling factor.
}
\label{fig:DUC}
\end{figure}


\subsubsection{Improved Shuffle Block}\label{sec:shuffleblock}
To obtain an effective lightweight network, we reduce the overall parameters while ensuring the number of valid parameters.
We introduce the blocks from ShuffleNetV2~\cite {ma2018shufflenet} and reduce the number of channels in the convolution operation to reduce the parameters and calculations.
To ensure the proportion of valid parameters in the network, we modify the shuffle blocks in ShuffleNetV2~\cite {ma2018shufflenet} and adopt an improved version to build the subnetwork in each stage.

The modified shuffle blocks are designed to accommodate the extremely lightweight network structure.
The detailed structures of the two types of improved shuffle blocks is shown in Fig.~\ref{fig:block}.
Each subnetwork in the four stages uses (a) as the first building block followed by several (b) blocks.
Different from the building blocks in ShuffleNetV2, we remove the rectified linear unit (ReLU)~\cite{glorot2011deep} after the first position-wise convolution layer in each shuffle block.
The ReLU activation function is defined as

\begin{equation}\label{Eq:relu}
ReLU\left( x \right) =
    \begin{cases}
        x & \text{if } x>0 \\ 
        0 & \text{if } x\le 0 
    \end{cases},
\end{equation}
where $x$ is the input of the ReLU.
For the negative inputs of ReLU, the outputs are all zeros.
This empowers the ReLU to effectively reduce the redundancy and the effect of over-fitting in large neural networks. 
However, the features of the light-weight network have already been compressed.
As a result, when ReLU collapses the channel of the feature map, it inevitably loses information in that channel.
In other words, some neurons will no longer be activated, resulting in a decrease in the number of active neurons.

The parameter amount of the light-weight network is limited.
Thus, inactivating some of the neurons within this network will attenuate its representation capacity.
To address this problem, we remove the ReLU after the first convolution layer in each shuffle block.
We keep the other ReLU in the shuffle block to maintain the non-linear capabilities of the network.
Therefore, the modified blocks have more active neurons making the light-weight network more efficient than the previous ones.

\subsubsection{Dense Upsampling Convolution}\label{sec:DUC}
Inspired by~\cite{wang2018understanding}, we adopt Dense Upsampling Convolution (DUC) to increase the resolution of feature maps and match the dimension across the adjacent subnetworks in the low-to-high process.
In~\cite {wang2018understanding}, DUC is used in the output layer to predict the label map as the final output of the network.
Whereas, in this paper, DUC is used for feature upsampling, which is a more lightweight approach compared to the conventional upsampling methods~\cite{chen2017deeplab,chen2017rethinking,chen2018encoder}.

Fig.~\ref{fig:DUC} depicts the architecture of the DUC layer.
The DUC operation is performed on an input feature map with height $h$, width $w$, and feature channels of $c$, and 
the output feature map has the dimension $ h \times w \times (d^2 * L)$.
Where $h = H/d$, $w = W/d$, and $d$ is the upsampling factor .
The output feature map is then reshaped to $H \times W \times L$.
Instead of performing bilinear upsampling, which is not learnable, or using a deconvolution network (as in~\cite{noh2015learning}), in which zeros have to be padded in the unpooling step before the convolution operation, DUC applies convolutional operations directly on the input feature map to get the upsampled feature map. 
It can be seen that DUC divides the entire output feature map into equal $ d ^ 2 $ sub-parts, which have the same height and width as the incoming feature map.
This division allows us to apply the convolution operation directly between the input and output feature maps without the need to insert extra values in upsampling networks, such as the deconvolutional networks or the unpooling operation.

\begin{figure}[!tb]
\centering
{
\includegraphics[width=0.6\linewidth]{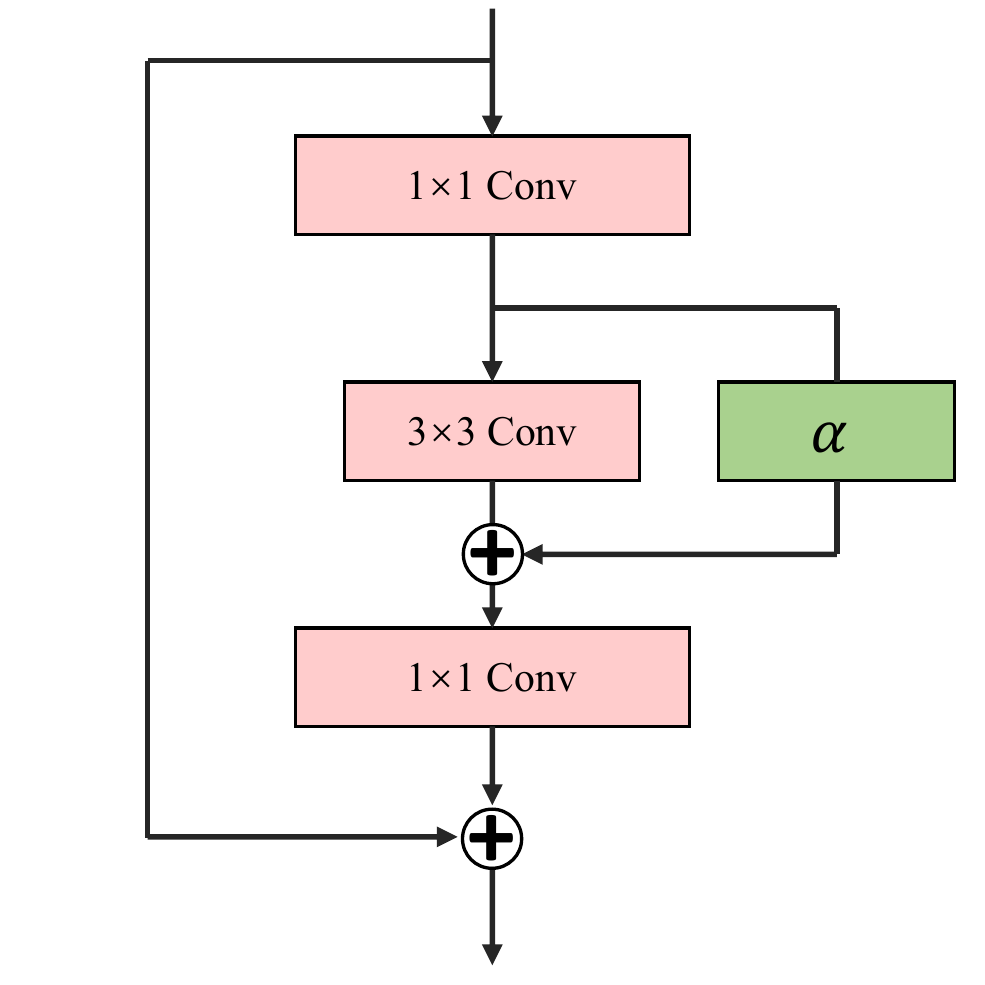}
}
\caption{
\textbf{Residual Adapter.} 
The figure shows a bottleneck residual module with the inclusion of adapter modules (bypass convolution with weights $\alpha$ in green).}
\label{fig:Adapter}
\end{figure}


\subsubsection{Residual Adapter}\label{sec:adapter}

The proposed method is a single-step approach to detect the keypoints and their association simultaneously, instead of detecting the bounding box of the UGV first and then performing keypoint detection like other two-step methods~\cite{wu2018simultaneous,papandreou2017towards}.
We use the heatmap to locate the keypoints and PAF to encode part-to-part association. 
Therefore, the output of the network is divided into two branches, which output heatmap and PAF, respectively.

There is an inherent gap between the two data distributions of heatmap and PAF.
The heatmap predicts the confidence of whether each pixel corresponds to a keypoint.
PAF reserves the direction vector field between each coadjacent keypoint to determine the part-to-part association.
In other words, the heatmap provides global positioning information, and the PAF provides the local relative relationship. 

However, heatmap and PAF use the same backbone to extract the feature maps and share the vast majority of their parameters.
We adopt the strategy in~\cite{rebuffi2018efficient} and use a bypass modular adapter to adjust the impact of data distribution through learning adaptively.
Concretely, in addition to the mainstream convolution $\mathcal{F}$, we also add a bypass convolution to learn the adapter $ \alpha $ with adaptation parameters $W_{ \alpha  }$ in parallel and then add it to the mainstream result.
The kernel size of the adapter is set to $1 \times 1$ to guide the features one by one to steer it to different target tasks employing a small number of parameters.
The adapter module has the form: 
\begin{equation}\label{eq:adapter}
    \mathbf{y}={ \mathcal{F} }\left\{  \mathbf{x}, W \right\} + W_{ \alpha  }{ \mathbf{x} },
\end{equation}
where $\mathbf{x}$ is the input feature map, $W$ is the parameter of the mainstream convolution $\mathcal{F}$, and $ \mathbf{y} $ is the output.

In our UGV-KPNet, the mainstream network of the module is a residual learning block with a shortcut connection, thus forming a residual adapter.
The structure of this residual adapter is shown in Fig.~\ref{fig:Adapter}.
The kernel size of mainstream convolution is $3\times 3$.
Through the point-wise convolution ($1\times 1$ conv) at both ends, the number of channels of the feature map is reduced and upgraded, so that the number of parameters and calculation can be effectively controlled.
Moreover, both the $1\times 1$ convolutions at two ends and the convolution in the middle adopt the depth-wise convolution~\cite{howard2017mobilenets,chollet2017xception} to make the network as lightweight as possible.
Therefore, there are very few parameters in this residual adapter.

\begin{figure}[!tb]
\centering
{
\includegraphics[width=0.95\linewidth]{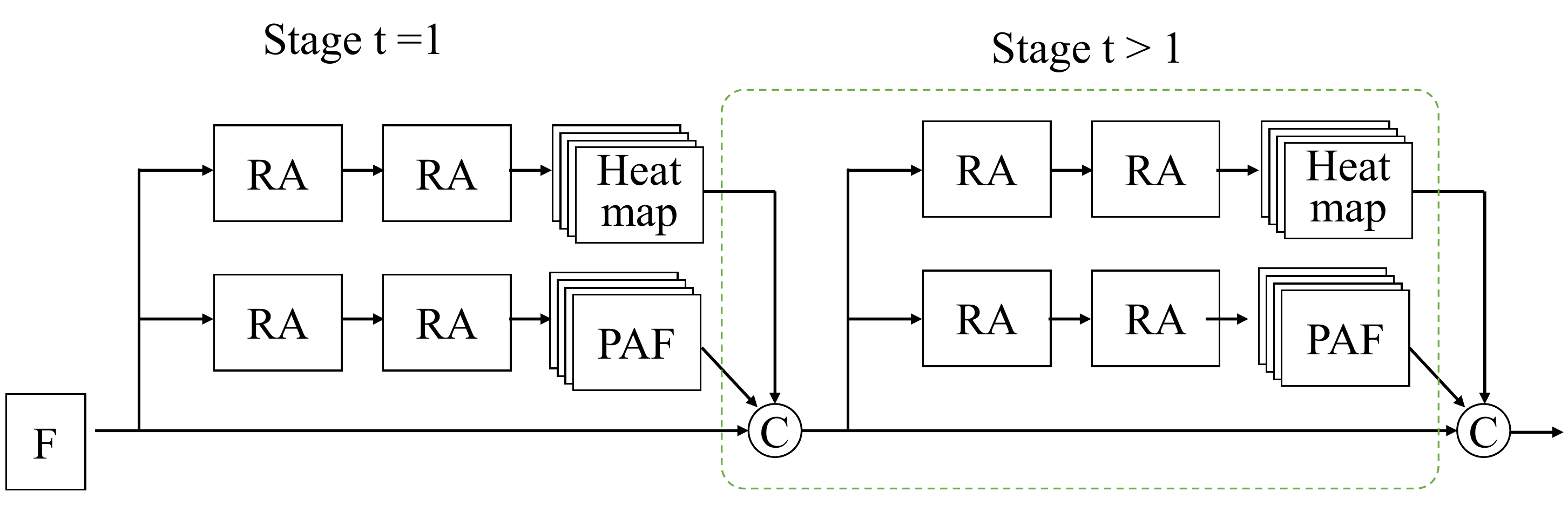}
}
\caption{
\textbf{Multi-stage Refinement.}
This module is a two-branch multi-stage CNN, where RA is the residual adapter block.
Each stage in the first branch predicts the heatmaps, and each stage in the second branch predicts the PAFs. 
After each stage, the predictions from the two branches, along with the image features F, are concatenated for next stage.
}
\label{fig:MultiStageRefinement}
\end{figure}

\subsubsection{Multi-stage Refinement}\label{sec:Multi_Stage_Refinement}
The prediction module has two branches to predict the heatmap and PAF, respectively.
Each branch is an iterative prediction architecture, following the approach in~\cite{OpenPoseTPAMI}, which refines the predictions over successive stages with intermediate supervision at each stage. 
Fig.~\ref{fig:MultiStageRefinement} shows the architecture of the multi-stage refinement.
At the first stage, the features F obtained from the backbone network are fed into two stacked residual adapters in each branch.
The predicted heatmap and part affinity fields are concatenated with F as the input for the next stage.

\subsection{Heatmap for the keypoints}\label{sec:Heatmap}

The proposed network predicts heatmap to locate the keypoints of the UGV.
Each channel of the heatmap denotes a confidence map, which is a 2D representation of the belief that a particular keypoint of the UGV occurs in the image.
The predicted heatmap of the network has five channels, four of which correspond to the keypoints of four categories, and an extra channel corresponds to the background.
Ideally, if a UGV appears in the image, a single peak should exist in each confidence map if the corresponding keypoint is visible. 
The background channel is a confidence map where each value at the corresponding location indicates whether it belongs to the background.

The ground truth heatmap is generated from the annotated keypoints in the 2D images.
Each value of the confidence map for all types of keypoint is initialized to zero.
In these confidence maps, pixels at positions corresponding to keypoints of a specified type have a confidence score.
The initial ground truth of the background channel is all ones.
We perform element-wise addition for the four keypoint channels, 
and then subtract the addition result from the initial value to obtain the ground truth of the background channel.

In previous methods (such as OpenPose~\cite{OpenPoseTPAMI}), the confidence score for each keypoint is obtained from a 2D Gaussian distribution mask, which is centered at the position of the keypoint and overlaid onto the initial confidence map.
The spread of the peak of the confidence score in each Gaussian mask is controlled to ensure spatial accuracy.
As a result, only a few points near the center of each mask have a high confidence score, while the remaining points have a deficient score.
When there are many keypoints in the image, since each keypoint has some high-scoring pixels in the confidence map, it can ensure a certain number of pixels with high confidence scores.
However, in our task, there is only one UGV with four keypoints in the image.
In the generated confidence map, the number of points with a high score is minimal, and most of them are background pixels with scores of zeros.
Therefore, using a Gaussian distribution to generate the ground truth will make the number of positive training points insufficient, which makes it difficult to train the network.
Another problem is that high-score points are too concentrated, which exacerbates the interference of labeling bias in the training dataset.
Keypoints manually labeled are sometimes few pixels away from their actual location.
Using a Gaussian mask with a centralized confidence score will cause deviations in the position of the peaks, 
leading to severe effects of manual labeling bias and unstable training.

To solve these two problems, we use a circular mask with binary scores to represent the confidence map of each keypoint to generate the ground truth in the training dataset.
On the one hand, all points in the circular binary mask have a high confidence score, which makes the number of positive points in the heatmap significantly increase compared to the Gaussian distribution.
On the other hand, when the labeled data is slightly biased, the pixels at the corresponding positions of the keypoints are still within the circular mask and have a high confidence score.

\begin{figure}[!tbp]
\centering
{
\includegraphics[width=0.8\linewidth]{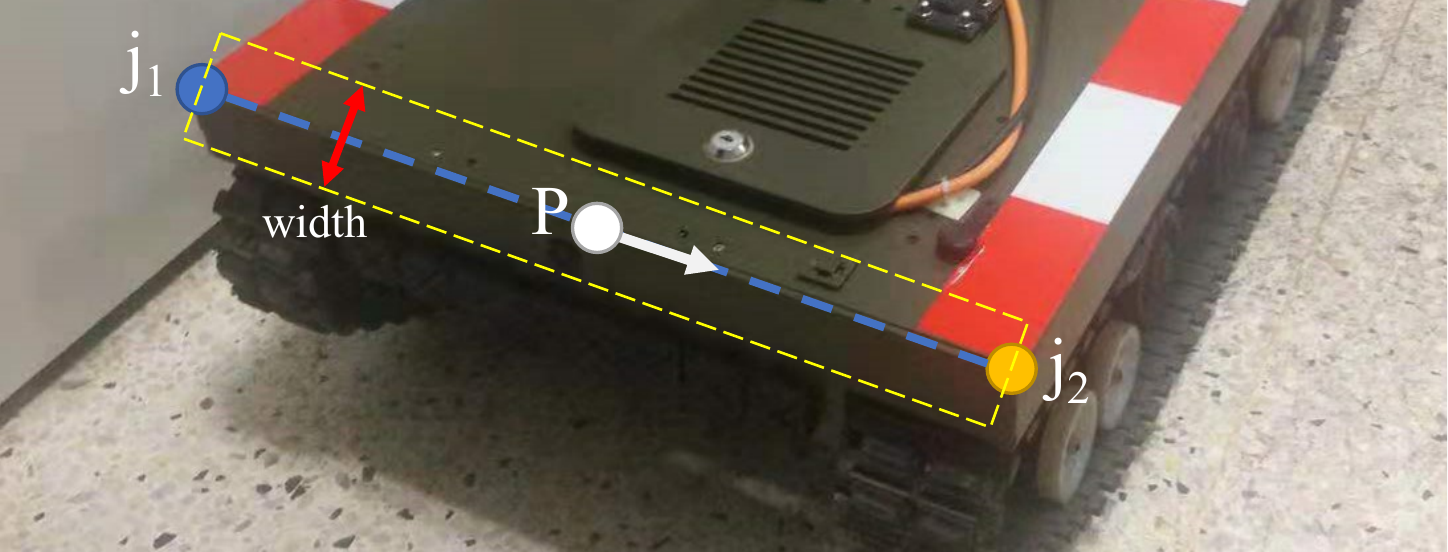}
}
\caption{
\textbf{PAF.}
The definition of PAF of the points lies on the connected edge of two keypoints. 
Suppose $\mathbf{p}$ is located on the connection of keypoints $j_1$ and $j2$, and its part affinity is a 2D unit vector point from $j_1$ to $j_2$.
}
\label{fig:PAF}
\end{figure}

\subsection{Part Affinity Fields for Part Association}\label{sec:PAF}
To determine the correlation between the keypoints, we adopt the part affinity field (PAF), which is first proposed in~\cite{cao2017realtime}.
PAF preserves both location and orientation information across the connected edge of two keypoints, as shown in Fig.~\ref{fig:PAF}. 
The part affinity is a 2D vector field for each connection. 
For each pixel in the area belonging to a particular connection, a 2D vector encodes the direction that points from one keypoint of the connection to the other. 
Each type of connection has a corresponding affinity field joining its two associated keypoints. 

The PAF is a two-channel feature map for the directed connection between two points.
As shown in Fig.~\ref{fig:PAF}, for a point P lying on the connection between two points, its part affinity is a 2D unit vector that encodes the direction from a keypoint $j_1$ to another keypoint $j_2$. 
The two components of this 2D unit vector are used as the ground truth at the corresponding positions in the two channels of the PAF.
For all other points that are not on the connection, the vectors are zero-valued, and their corresponding ground truth values are zeros in the PAF.
The connection between two keypoints is a line segment with a certain width.
The set of points on the connection is defined as points within a distance from the line segment.
This distance threshold is half the width of the connection.

When there are multiple UGVs, PAF can be used to determine the connection relationship of keypoints. 
When there is only one UGV, PAF can effectively reduce false keypoint detection. 
If a predicted keypoint is a false detection, no PAF is connected to it.
Therefore, the prediction result of the PAF can be used to screen whether the predicted keypoint is a false positive comprehensively.
In the wild and other complex environments, the surrounding background may be very close to the texture of the UGV, resulting in some false positives of keypoints predicted by the network.
In this case, PAF can reduce the interference of false detections on the final result.

In the proposed approach, each keypoint is connected to the next one, and the last connected to the first. 
The connection relationship between the four keypoints is shown in Fig.~\ref{fig:def_keypoints}.
The four connection edges then form a quadrilateral, providing an 8-channel ground truth PAF.
We increase the connection width from 1 pixel to 6 pixels to improve the stability and robustness of the network.

\subsection{Assemble}\label{sec:assemble}

During prediction, multiple predictions with different confidence scores may appear around the position of a keypoint. 
We perform non-maximum suppression (NMS) on the confidence maps to obtain a discrete set of keypoint candidates.
For each keypoint, we may have more than one candidate due to false positive predictions.
Then, to get the ordered structure of the UGV,
we employ the assembling strategy of~\cite{OpenPoseTPAMI} to deal with the matching problem of the connection candidates and keypoint candidates.
During the assembling, some false positive keypoints predictions can be eliminated through the relationship between the keypoints and PAF.

\subsection{Implementation Details}
   
\subsubsection{Loss function}
During training, we use the mean squared error (MSE) loss, which creates a criterion that measures the discrepancy between the ground truth and the prediction for both the heatmap and the PAF.
The training loss $\mathcal{L}$ is defined as:
\begin{equation}\label{eq:loss}
\mathcal{L}=\frac { { w }_{ p } }{ ns } \sum _{ j }^{ s }{ \sum _{ i }^{ n }{ { \left( { p }_{ i }^{ j }-{ \hat { p }  }_{ i }^{ j } \right)  }^{ 2 } }  } +\frac { { w }_{ q } }{ nt } \sum _{ j }^{ t }{ \sum _{ i }^{ n }{ { \left( { q }_{ i }^{ j }-{ \hat { q }  }_{ i }^{ j } \right)  }^{ 2 } }  } ,
\end{equation}
where ${ p }_{ i }^{ j }$ is the prediction of the $i$th element in $j$th channel of the heatmap, and ${ \hat { p }  }_{ i }^{ j }$ is the ground truth. Similarly,
${ q }_{ i }^{ j }$ is the prediction of the $i$th element in $j$th channel of the PAF, and ${ \hat { q }  }_{ i }^{ j }$ is the  ground truth.
${ w }_{ p }$ and ${ w }_{ q }$ are the balancing weights for the two loss terms, which are set to $1$ and $0.1$ in the experiments.

\subsubsection{Data balancing}
Since most of the regions in the image are background, only a few areas correspond to the keypoints and connected edges of the UGV.
To deal with this imbalanced data distribution,
we sample the background area in both the heatmap and PAF of the training data. 
Only the sampled area is involved in the calculation of loss, and the other regions are ignored during the training process.
We set the initial sampling rate to 5\%, and then increase it by 2\% per training epoch until all regions are used in training.

\subsubsection{Training Protocol}
Given the training dataset, the proposed network can be trained end-to-end.
All the experiments are conducted using the PyTorch framework on GPUs.
The input images are resized to $360 \times 640$ during training and inference.
The model is trained using the SGD optimizer with a momentum of 0.9, weight decay of $10^{-4}$.
We train the network for 400 epochs with a batch size of 24.
The initial learning rate is 0.5, and we reduce the learning rate by a factor of 2 when the metric loss improves less than 1e-8 within five consecutive epochs.
Aiming at solving the problem that the vision-based method is susceptible to light changes, we use a data augmentation method to ensure the stability of detection.
During training, we randomly change the brightness and contrast of the image with a certain probability when processing the input data.


\section{6-DoF pose estimation}\label{sec:6dof}
In this section, we introduce the method that we used to estimate the 6-DoF pose of the UGV in a world reference frame $\mathbf{F}^{W}$. 
We set a robot (UGV) reference system $\mathbf{F}^{R}$  with the UGV's center as the origin. According to the known UGV dimension information, a set of 3D coordinates $\Phi $ in $\mathbf{F}^{R}$ that correspond to the four 2D coordinates $\varphi$ of the detected keypoints in the 2D image can be obtained.
Then, based on 3D coordinates $\Phi $, 2D coordinates $\varphi$, and the calibrated intrinsic parameters of the camera, the 6-DoF pose of the robot reference system $\mathbf{F}^{R}$ relative to the camera reference frame $\mathbf{F}^{C}$ can be obtained as ${ k }_{ r }^{ c }$.
The pose of the camera reference frame $\mathbf{F}^{C}$ relative to the world reference frame $\mathbf{F}^{W}$ can be obtained as ${ k }_{ c }^{ w }$ in a similar way by setting the UGV at the origin of the world reference frame.
The pose of UGV in the world reference frame ${ k }_{ r }^{ w }$ can be obtained by converting  the ${ k }_{ r }^{ c }$ according to ${ k }_{ c }^{ w }$.  

\noindent
\textbf{Robot reference frame.}
First, we set up the robot reference frame $\mathbf{F}^{R}$.
The center of the UGV is aligned as the origin of the robot reference frame.
Based on the orientation of the UGV, the X-axis is the starboard, the Y-axis is the forward, and the Z-axis is the zenith.
The 3D coordinates of the four keypoints in the $\mathbf{F}^{R}$ can be easily obtained according to the pre-measured dimensions $L \times W \times H$ of the UGV,
where $L$ is the length of the UGV, $W$ is the width, and $H$ is the height.
Four key points are located at the four vertices of the upper surface of the UGV, and their corresponding coordinates $\Phi $ are 
$(-\frac{W}{2},\frac{L}{2},\frac{H}{2})$,
$(-\frac{W}{2},-\frac{L}{2},\frac{H}{2})$,
$(\frac{W}{2},-\frac{L}{2},\frac{H}{2})$, 
and $(\frac{W}{2},\frac{L}{2},\frac{H}{2})$, respectively.

\begin{figure*}[!htbp]
\centering
{
\includegraphics[width=1.0\linewidth]{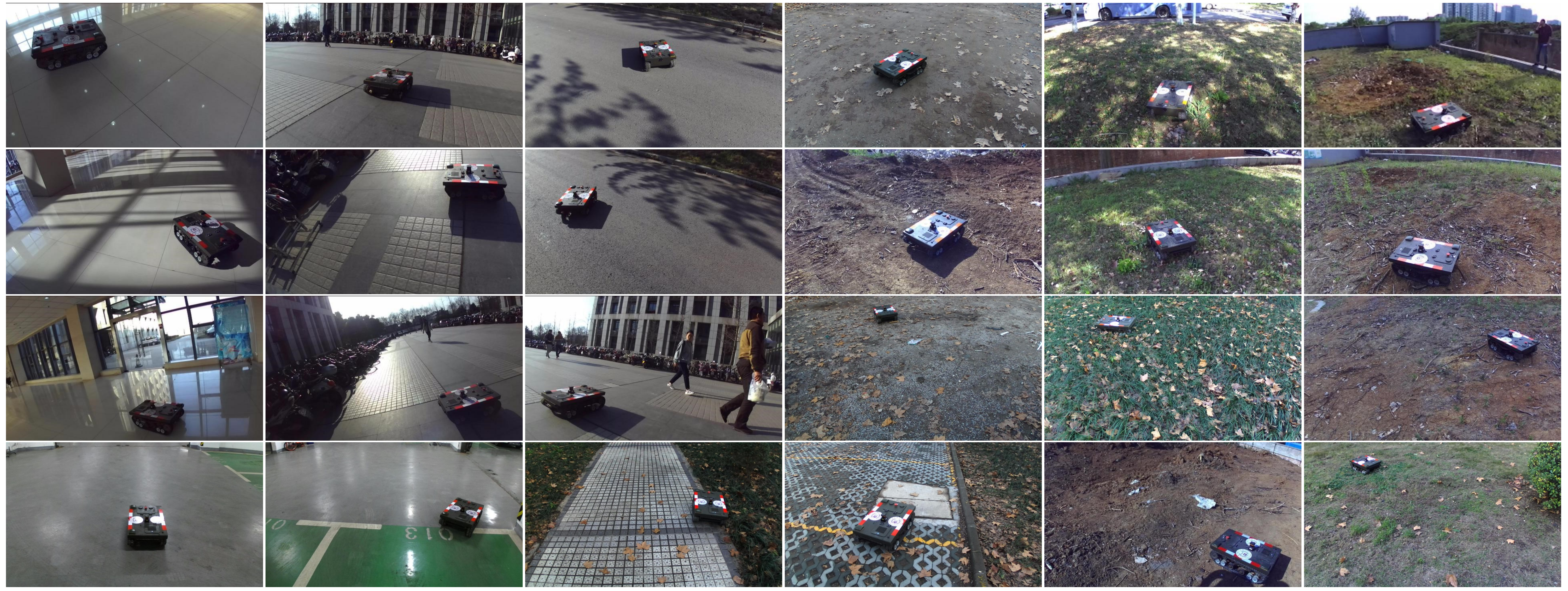}
}
\caption{\textbf{UGVKP dataset.} We collected a large-scale UGV keypoints dataset to train our network. For each of the scenes, we select a set of camera positions to capture several images. 
All the images are accurately labelled the key points of the UGV.}
\label{fig:dataset}
\end{figure*}

\noindent
\textbf{Camera reference frame.}
Second, we calculate the transformation relationship from $\mathbf{F}^{R}$ to the camera reference frame ($\mathbf{F}^{C}$).
This is a Perspective-n-Point (PnP) problem, in particular, a P4P problem. 
Each keypoint provides a 3D-2D point correspondence.
Since the four keypoints are co-planar, we adopt Levenberg-Marquardt optimization~\cite{roweis1996levenberg} which is an iterative method to solve this P4P problem.
EPnP~\cite{lepetit2009epnp} is another classic method for solving the PnP problem, and has higher efficiency than Levenberg-Marquardt optimization.
However, Levenberg-Marquardt optimization gives significantly better results than EPnP.
Since there are only four corresponding points in our PnP problem, the calculation time required by these two methods is almost the same.
By solving the PnP problem, we can obtain the rotation matrix $R$ and the translation vector $T$.

\noindent
\textbf{World reference frame.}
Third, we calculate the pose of the $\mathbf{F}^{C}$ relative to a world reference frame ($\mathbf{F}^{W}$).
Let the origin of the world reference frame be at the center of the UGV before the UGV leaving the loading mechanism.
At this moment, the robot reference frame coincides with the world reference frame.
The translation vector $T_{0}$ and rotation matrix $R_{0}$ in the current state can be calculated as above.
The pose of the $\mathbf{F}^{W}$ relative to the $\mathbf{F}^{C}$ equals to the $\mathbf{F}^{R}$ relative to the $\mathbf{F}^{C}$.
The translation vector $T_{c}$ and rotation matrix $R_{c}$ represent the pose of the $\mathbf{F}^{C}$ to the $\mathbf{F}^{W}$ can be obtained with the rotation matrix $R_{0}$ and translation vector $T_{0}$ as below.

\begin{equation}\label{eq:camera2UGVR}
R_{c}=R_{0}',
\end{equation}
\begin{equation}\label{eq:3dcoordinate}
T_{c}=-R_{0}'\cdot T_{0}.
\end{equation}

Last, to get the pose of UGV under the world reference frame, we combine the UGV to camera pose with the camera to world pose by

\begin{equation}\label{eq:camera2worldR}
R_{w}=R_{c}\cdot R,
\end{equation}
\begin{equation}\label{eq:camera2worldT}
T_{w}=T_{c}+R_{c}\cdot T.
\end{equation}

Where
$R_w$ and $T_{w}$ are the UGV's rotation matrix and translation vector in the world reference system.
That is,  $T_{w}$ is the 3D coordinate of the UGV in the world reference system.
The Euler angles of the UGV can be calculated from the elements of $R_{w}$ as below.

\begin{equation}\label{eq:eulerz}
\theta _{z}=\arctan \left( r_{21},r_{22}\right),
\end{equation}
\begin{equation}\label{eq:eulerx}
\theta _{x}=\arctan \left( r_{13},r_{33}\right),
\end{equation}
\begin{equation}\label{eq:eulery}
\theta _{y}=\arcsin \left(-r_{23}\right),
\end{equation}
where $r_{ij}$ represents the element in $i$-th row and $j$-th coloumn of $R_{w}$.
The rotation order is Z-X-Y, which corresponds to the yaw-pitch-roll from the perspective of the UGV.
The final output of the 6-DoF pose estimation is the 3D position $T_w$ and the attitude 
$\left( \theta _{z}, \theta _{x}, \theta _{y} \right)$.

\section{Training Data for UGV Keypoints Detection}
One of the main obstacles for training the deep networks for keypoints detection is the lack of large-scale annotated datasets, especially for the keypoint detection task of UGV.
In this paper, we create a real-world dataset tailored for UGV keypoints, which is called UGVKP, containing both challenging indoor and outdoor environments. The dataset includes a lot of locations and poses variations to ensure the diversity of the images.

%
%
\begin{figure}[!htbp]
\centering
{
\includegraphics[width=0.75\linewidth]{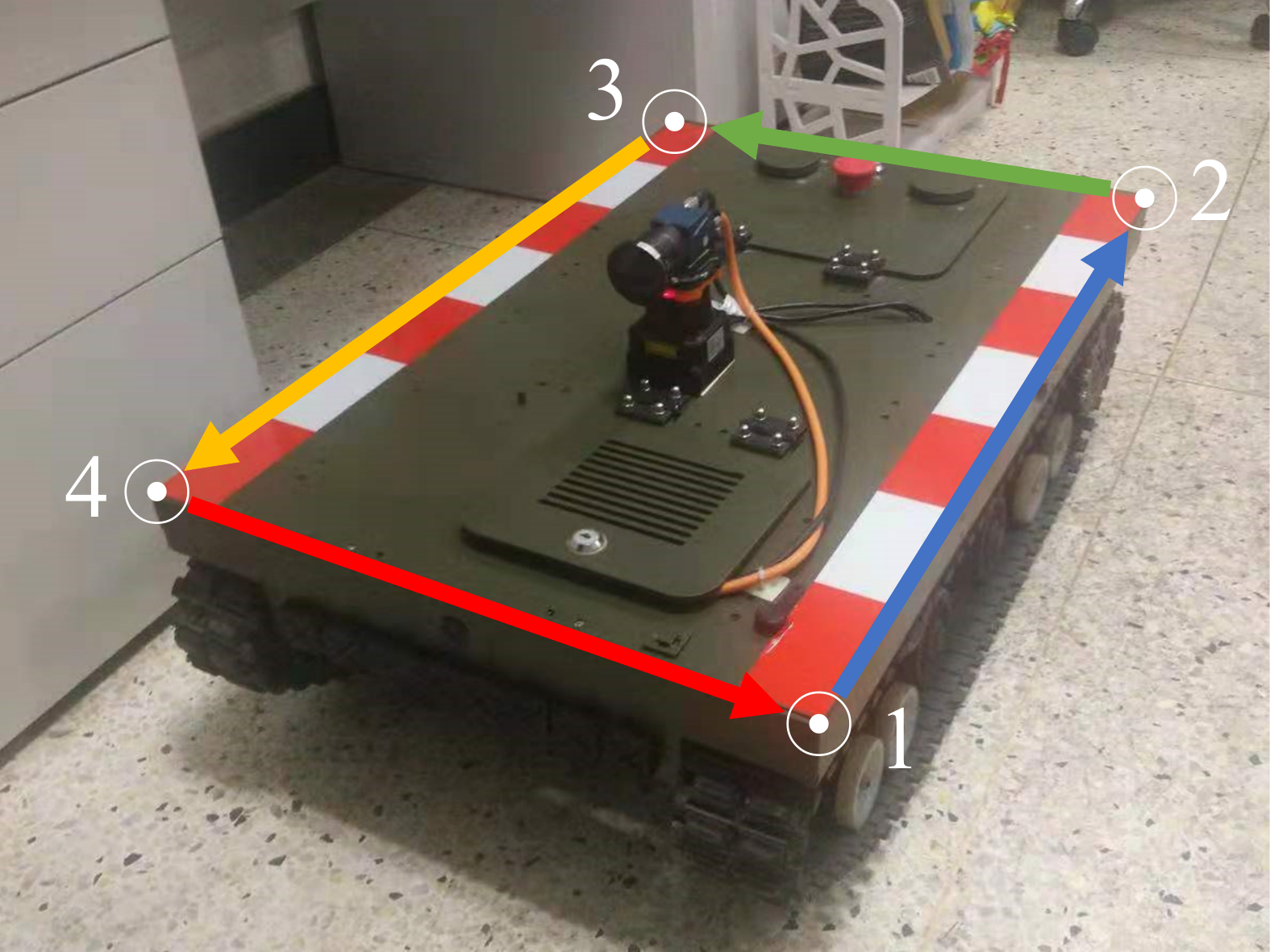}
}
\caption{The definition of the UGV keypoints and connections.}
\label{fig:def_keypoints}
\end{figure}

\subsection{UGVKP: UGV keypoints dataset}

The UGVKP dataset contains 894 different images of the UGV, and these images are randomly divided into 75\% as the training set and the other 25\% as the test set. The images are with a size of $1080 \times 1920$.
UGVKP includes various outdoor scenes such as grass, mud, urban roads, stone roads, and other outdoor environments, as well as indoor scenes such as parking lots, indoor stadiums, etc.
In each scene, we adjust the pose and position of UGV in two ways to collect images.
One way is to capture continuous images of the UGV driving from far to near w.r.t the camera from various directions.
The other way is to place the UGV at different positions with various poses randomly.
The images in UGVKP are also captured under various lighting conditions such as shadows, intense lighting, dim lighting, etc.
Fig.~\ref{fig:dataset} shows example images from the UGVKP dataset. For each image, the ground truth keypoints are annotated manually. 

\subsection{Keypoints and Connections Definitions}\label{sec:definition}
The definition of the keypoints of a UGV is shown in Fig.~\ref{fig:def_keypoints}.
Each UGV has four keypoints, the bottom right is the first point, the back right is the second, then the top left is the third, and the bottom left is the last.
We label the position of each keypoint in turn.
As shown in Fig.~\ref{fig:def_keypoints}, the connection between two keypoints is shown by the arrow pointing from one keypoint to the other. Different connections are marked in different colors.


\begin{table*}[!htbp]
\caption{
\textbf{Comparison to the state-of-the-art.}
The number of parameters, FLOPs, Precision, Recall, IoU and average distance of our posed UGVPoseNet and comparison with other state-of-the-art methods on the validation set.
UGVPoseNet* is an extremely lightweight version with very few parameters and calculations, which has only one resolution subnetworks in the backbone.
}
\begin{center}
\scalebox{1.0}
{
\begin{tabular}{c|l|c|c|c|c|c|c}
\hline

Framework                   
    & Backbone
    & Params (M) 
    & FLOPs (G)
    & Precision (\%) 
    & Recall (\%)
    & IoU (\%) 
    & Dist (pixels) \\ \hline
OpenPose~\cite{OpenPoseTPAMI}    & VGG19         & 49.5  & 221.4         & 92.66 & 89.09 & 92.54 & 2.789 \\ \hline  

Ours        & Xception~\cite{chollet2017xception}      &  20.8 &   78.7        & 92.12  & 89.75 & 92.17 & 3.68 \\   

Ours        & ResNet18~\cite{he2016deep}      &  11.2 &   42.4        & 88.25  & 85.46 & 91.09 & 3.63 \\   

Ours        & MobileNetV2~\cite{sandler2018mobilenetv2}   &  2.24 &   8.5         & 91.33  & 89.77 & 92.11 & 3.61 \\   

Ours        & ShuffleNetV2~\cite{ma2018shufflenet}  &  1.27 &   6.19        & 91.56  & 90.16 & 91.86 & 3.84 \\  

Ours        & UGVPoseNet*       &  \textbf{0.33} &   \textbf{1.25}        & 89.73  & 86.75 & 93.40 & 2.38 \\ \hline  

Ours        & UGVPoseNet                  &  2.93 &   2.68        & \textbf{93.33}  & \textbf{92.72} & \textbf{93.68} & \textbf{2.20} \\ \hline  

\end{tabular}
}
\label{tab:SOTA}
\end{center}
\end{table*}

\section{Experiments}\label{sec:Experiment}
We validate the proposed solution on a combination of a UGV and a carrier vehicle. 
As shown in Fig.~\ref{fig:def_keypoints}, the UGV is a self-designed crawler ground mobile vehicle.
Its dimensions (length, width, and height) are $72 cm \times 48 cm \times 23 cm$.
The carrier vehicle is a modified off-road vehicle.
The dimensions (length, width, and height) of the hanging frame mounted on the carrying vehicle are $90cm \times 65cm \times 5cm$.
The composition of the entire system is shown in Fig.~\ref{fig:teaser}.
In this section, we focus on the algorithm part of the recovery system and conduct a series of experiments to verify the keypoints detection and 6-DoF estimation.

\subsection{Evaluation metrics}
The proposed method combines the heatmap of the keypoints and PAF of the connected edges with an assembling strategy to detect and locate the UGV and then estimates the 6-DoF pose of the UGV.
To evaluate the method effectively and comprehensively, we set some metrics to assess the algorithm from the two aspects of detection and positioning quantitatively.

In terms of detection, we consider the UGV with four keypoints as a whole to evaluate the detection effect of the proposed network.
We use the quadrilateral structure assembled from the predicted keypoints as the bounding box of a UGV, and calculate the pixel-level intersection over union (IoU) between the prediction and the ground truth.
We calculate the average IoU of all predicted UGVs and the ground truth as the overall evaluation metric for keypoints detection.
Besides, we calculate the precision and recall to evaluate the detection results.
We set the prediction with the IoU value higher than a threshold $T$ as True Positive (TP), otherwise as False Positive (FP), and those that are not detected as False Negative (FN).
To obtain reliable keypoints accurate pose estimation, we set $T$ to be a relatively large value of $80\%$.
When a UGV is detected multiple times, we take the prediction with the highest IoU as TP and the rest as FP. 
Then, we can get $ Precision = TP / {(TP + FP)} $ and $ Recall = TP / (TP + FN) $.


To evaluate the positioning accuracy of each keypoint, we calculate the Euclidean distance of the predicted keypoints in the 2D images with their ground truth.
We evaluate each type of keypoint separately and get the corresponding errors. 
The average error of the four types of keypoints is used as the overall evaluation for keypoint positioning.

The 6-DoF pose estimation is obtained by solving the PnP problem based on the predicted keypoint results.
Then, the vertexes of the UGV is projected to the image using the estimated 6-DoF pose. 
Each vertex corresponds to a projected keypoint.
The comparison between the projected keypoints and the ground truth keypoints is conducted to evaluate the system accuracy of our proposed UGV position and pose estimation approach.

\subsection{Comparison to alternative approaches}

The framework of the proposed keypoint detection method is a practical improvement based on the state-of-the-art method OpenPose~\cite{OpenPoseTPAMI}.
Therefore, we first compare the performance of our method with OpenPose.
OpenPose adopts VGG19 as the backbone network and achieves excellent precision and recall. 
However, its parameters and computational costs are much higher than our method.
As shown in Table.~\ref{tab:SOTA}, OpenPose has 49.5M parameters and 221.4G FLOPs, whereas our method has only 2.93M parameters and 2.68G FLOPs while achieving 0.67\% higher precision and 3.63\% higher recall than OpenPose.

We also compare the proposed lightweight backbone network with several state-of-the-art backbone networks under the same framework.
In particular, we replace the backbone network with Xception~\cite{chollet2017xception}, ResNet18~\cite{he2016deep}, MobibleNetV2~\cite{sandler2018mobilenetv2} and ShuffleNetV2~\cite{ma2018shufflenet}.
Among these methods, UGV-KPNet achieves the highest keypoints detection accuracy, and it has the least computational cost, which is a more plausible choice for real applications on vehicle-mounted platforms.
Besides, we present an extremely lightweight version denoted as UGV-KPNet* in Table~\ref{tab:SOTA} with very few parameters and calculations. 
Compared with UGV-KPNet, UGV-KPNet* has only one resolution rather than four different resolutions in the subnetworks of the backbone. It only requires 0.33M parameters and 1.25 FLOPs. UGV-KPNet* performs slightly worse than UGV-KPNet, and it is suitable for situations with minimal computing resources.


\subsection{Qualitative Analysis}

\subsubsection{Visualization of the keypoints detection}

\begin{figure*}[ht]
\centering
{
\includegraphics[width=0.98\linewidth]{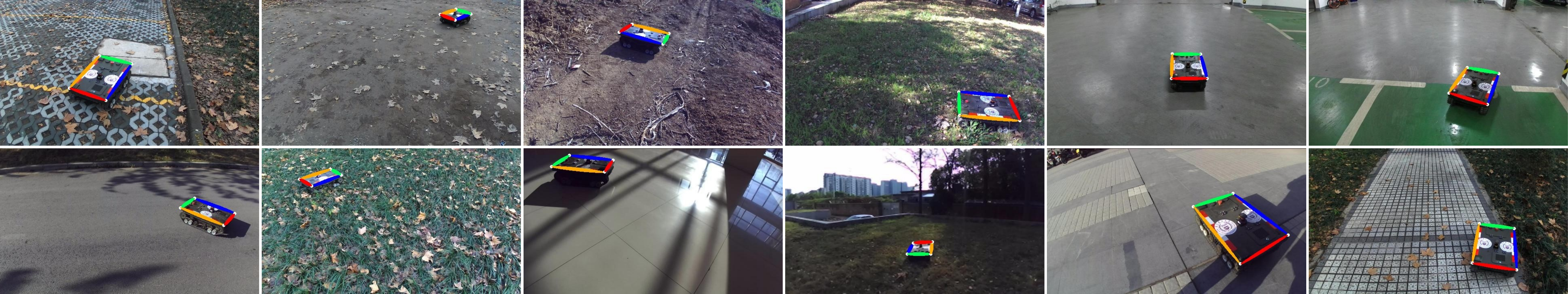}
}
\caption{
\textbf{Visualization of the keypoints detection.}
Different types of keypoints are connected by edges with different colors.
}
\label{fig:vis_keypoints}
\end{figure*}

Fig.~\ref{fig:vis_keypoints} shows some qualitative results of the proposed UGV-KPNet for keypoints detection. 
In the visualization, different types of keypoints are connected by edges with different colors.
UGV-KPNet has achieved promising detection results in both indoor and outdoor scenes, and can easily cope with changes in lighting and complex backgrounds.
The proposed UGV-KPNet can effectively detect the keypoints of the UGV with various attitudes at different distances. It can be seen that the predicted keypoints almost coincide with the four vertices of the UGV, indicating the high accuracy of the proposed approach.


%
%
\begin{figure*}[ht]
\centering
{
\includegraphics[width=0.98\linewidth]{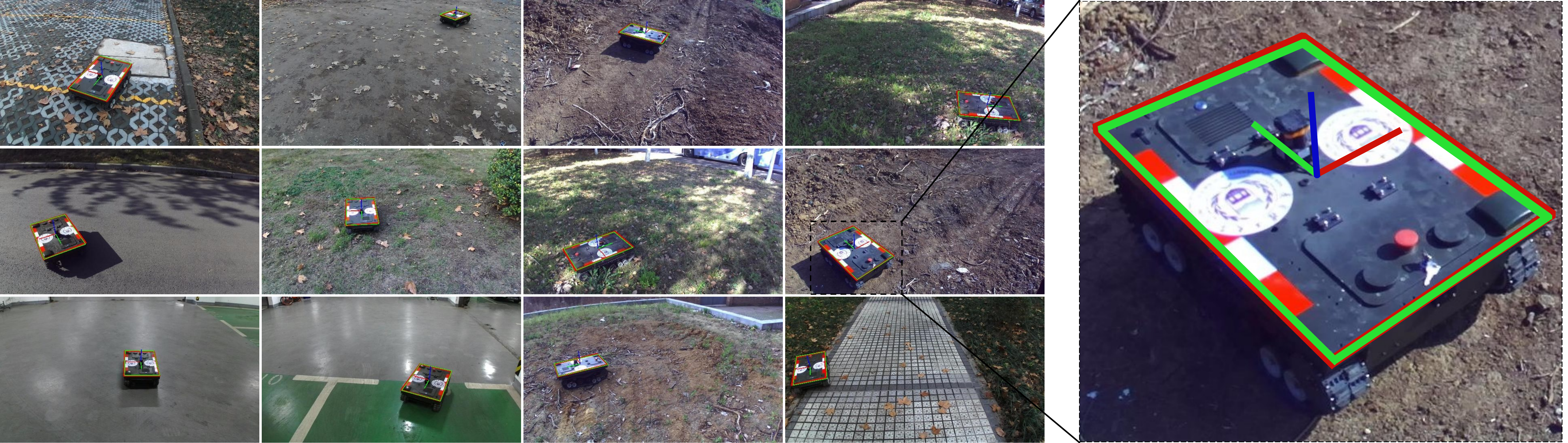}
}
\caption{
\textbf{Visualization of 6-DoF estimation.} 
The red bounding box is the connection of the four ground truth keypoints. The green box is the connection of the back-calculated keypoints based on the pose and 3D size of the UGV.
The red, green, and blue lines located in the center of UGV represent the three axes of the robot reference system.
}
\label{fig:vis_6DOF}
\end{figure*}

\subsubsection{Visualization of the 6-DoF pose estimation}

In Fig.~\ref{fig:vis_6DOF}, we visualize the results of the 6-DoF estimation. We draw the robot reference frame and keypoints calculated by the 6-DoF pose on the image.
The red, green, and blue lines located in the center of UGV represent the three axes of the robot reference system. 
The vertices of red and green boxes on the upper surface of the UGV represent keypoints of the UGV. The red ones are the ground truth, and the green ones are the estimation of the proposed method.
As shown in Fig.~\ref{fig:vis_6DOF}, our prediction and the ground truth almost overlap with each other, showing the superior accuracy of our approach.

\subsection{Ablation Study}
To verify the effectiveness of each of the improvements in the proposed method, we conduct an ablation study to analyze each component of the approach.

\subsubsection{Multi-resolution subnetworks}
To validate the effectiveness of the multi-resolution subnetworks, we remove this feature from UGV-KPNet.
Specifically, 
we keep only the first block with high resolution at each stage and remove the other blocks with different resolutions from UGV-KPNet, while the short-connections from the four stages are kept. We denote the resulting model as UGV-KPNet*, which is extremely light-weight compared with UGV-KPNet. UGV-KPNet* achieves slightly worse results than UGV-KPNet, but the results are still considerably well. As can be seen in Table~\ref{tab:SOTA}, UGV-KPNet* performs better than the state-of-the-art OpenPose as well as other counterparts of our methods with different backbones. The precision and recall are 89.73\% and 86.75\% for UGV-KPNet*. 
However, adding low-to-high and high-to-low network structures to UGV-KPNet* will increase the precision and recall by 3.6\% and 5.97\%, respectively.
The low-to-high and high-to-low structure has played an essential role in improving the accuracy of the network.
It can be seen that UGV-KPNet* has great advantages in terms of parameter number and computational cost.
It is suitable for situations where computing resources are very limited.
Or it can be applied to simple scenes with a clean background.
To obtain stable positioning and pose estimation of the UGV in various scenarios, the network with low-to-high and high-to-low network structure is the best trade-off.

\subsubsection{Improved Shuffle Block}

\begin{table*}[ht]
\caption{
Ablation results of the \textbf{Improved Shuffle Block} on our validation set. '-w/o ISB' indicates the network using the original shuffle block in \cite{ma2018shufflenet}.
The proposed UGVPoseNet and UGVPoseNet* are superior to the original versions in accuracy indicators respectively.
}
\begin{center}
\scalebox{1.0}
{
\begin{tabular}{l|c|c|c|c|c|c}
\hline
Methods     & Params (M)    & Flops (G)    & Precision (\%)    & Recall (\%)  & IoU  & Dist (pixels)\\ \hline     
UGVPoseNet-w/o ISB                &  2.93 & 2.68 & 90.85 & 92.05 & 93.01 & 2.38 \\  
UGVPoseNet                      &  2.93 & 2.68 & \textbf{93.33} & \textbf{92.72} & \textbf{93.68} & \textbf{2.20} \\ \hline  
UGVPoseNet*-w/o ISB &  0.33 & 1.25 & 86.39 & 84.11 & 91.98 & 3.58\\
UGVPoseNet*            &  0.33 & 1.25 & \textbf{89.73} & \textbf{86.75} & \textbf{93.40} & \textbf{2.38} \\ \hline  
\end{tabular}
}
\label{tab:abl_improved-shuffle-block}
\end{center}
\end{table*}

To verify the effectiveness of the proposed improved shuffle block, we designed two controlled experiments. Concretely, we replace the improved Shuffle Block in UGV-KPNet and UGV-KPNet* with the vanilla block of~\cite{ma2018shufflenet}, and compare the results against UGV-KPNet and UGV-KPNet*. The two modified counterparts of UGV-KPNet and UGV-KPNet* are denoted as UGV-KPNet-w/o ISB and UGV-KPNet*-w/o ISB. The results are shown in  Table~\ref{tab:abl_improved-shuffle-block}.
We can see that UGV-KPNet and UGV-KPNet* with the improved block has better performance on all evaluation metrics than UGV-KPNet-w/o ISB and UGV-KPNet*-w/o ISB. 
UGV-KPNet has 2.48\% and 0.67\% higher precision and recall than UGV-KPNet-w/o ISB.
The distance error of UGV-KPNet is 0.18 pixels smaller than UGV-KPNet-w/o ISB.
The advantages of the improved method are more evident for UGV-KPNet*.
It can be seen that UGV-KPNet* has 3.34\% and 2.64\% higher precision and recall than UGV-KPNet*-w/o ISB.
The distance error of UGV-KPNet* is 1.2 pixels smaller than UGV-KPNet*-w/o ISB.
The experimental results have clearly shown the effectiveness of our improved shuffle block.
Meanwhile, in both UGV-KPNet and UGV-KPNet*, the number of additional parameters introduced by the improved shuffle block is negligible.

\subsubsection{Multi-level connection}


\begin{table*}[!tb]
\centering
\caption{Ablation results of the \textbf{Multi-level Connection} and the \textbf{Residual Adapter} on our validation set.}
\scalebox{1.0}
{
\begin{tabular}{l|c|c|c|c|c|c}
\hline
Methods     & Params (M)    & Flops (G)    & Precision (\%)    & Recall (\%)  & IoU  & Dist (pixels)\\ \hline

UGVPoseNet-w/o MC   &  2.93     &   2.57    & 89.93  & 88.74 & 93.52 & 2.49 \\

UGVPoseNet-w/o RA   &  2.93     &   2.68    & 92.67  & 92.05 & 93.13 & 2.98 \\  

UGVPoseNet          &  2.93     &   2.68 & \textbf{93.33}  & \textbf{92.72} & \textbf{93.68} & \textbf{2.20} \\ \hline  
\end{tabular}
}
\label{tab:abl_mlc_ra}
\end{table*}

To verify the effectiveness of the multi-level connection, we experiment with a variant of the proposed model by removing the multi-level connection while maintaining the rest of the model. We denote the variant as UGV-KPNet-w/o MC. The results are shown in Table~\ref{tab:abl_mlc_ra}.
Removing the multi-level connection causes 3.4\% and 3.98\% decrease for precision and recall, respectively.

The performance gain introduced by the multi-level connection is two-fold.
On the one hand, multi-level information can better adapt to different sizes of the UGV in the images.
In the images with a wide field of view, the scale of the UGV varies at different positions.
The information from different stages has different receptive fields, which correspond to different scales. 
This multi-level structure enables the network to adaptively select information for the UGV in the images with different scales.
In Table~\ref{tab:abl_mlc_ra}, the recall with the multi-level connection structure is higher than that without it, 
which indicates that fewer keypoints are missed by our proposed method.
On the other hand, the multi-level structure provides both low-level texture features and high-level semantic features for the network to predict accurate keypoint detection in scenes with complex backgrounds. 
Semantic information can help the network determine the categories of the keypoints, and texture features can provide more accurate positioning information.
As can be seen in Table~\ref{tab:abl_mlc_ra}, UGV-KPNet obtains higher precision than UGV-KPNet-w/o MC,
which indicates that UGV-KPNet incorrectly detected fewer keypoints.

Also, when the multi-level connection is adopted, the average positioning error of each keypoint is reduced from 2.49 pixels to 2.20 pixels.


\subsubsection{Residual Adapter}
To prove its effectiveness, we experiment with a model by removing the residual adapter from our UGV-KPNet. We denote this variant as UGV-KPNet-w/o RA. The results are reported in Table~\ref{tab:abl_mlc_ra}. 
Compared with UGV-KPNet-w/o RA, we can see that our UGV-KPNet improves the precision from 92.67\% to 93.33\% and the recall from 92.05\% to 92.72\%.
The average distance error of keypoints has been reduced from 2.98 to 2.20.
Compared with all the parameters UGV-KPNet, this residual adapter only introduces a negligible amount of additional parameters, but it has a significant contribution to improving accuracy.
Therefore, this structure is very cost-effective and is suitable for the design of lightweight networks.

\begin{table}[!tb]
\caption{
The inference speed, precision and recall of different methods.
}
\begin{center}
\scalebox{1.0}
{
\begin{tabular}{c|l|c|c|c}
\hline
Framework                   
    & \multicolumn{1}{c|}{Backbone} 
    & \begin{tabular}[c]{@{}c@{}}Speed\\ (FPS)\end{tabular} 
    & \begin{tabular}[c]{@{}c@{}}Prec.\\ (\%)\end{tabular} 
    & \begin{tabular}[c]{@{}c@{}}Recall\\ (\%)\end{tabular} 
\\ \hline
OpenPose~\cite{OpenPoseTPAMI}    & VGG19         & 2.01  & 92.66 & 89.09 \\ \hline  
Ours        & Xception~\cite{chollet2017xception}      & 7.85  & 92.12  & 89.75 \\   
Ours        & ResNet18~\cite{he2016deep}      & 8.10  & 88.25  & 85.46 \\   
Ours        & MobileNetV2~\cite{sandler2018mobilenetv2}   & 19.81 & 91.33  & 89.77 \\   
Ours        & ShuffleNetV2~\cite{ma2018shufflenet}  & 19.47 & 91.56  & 90.16 \\ \hline  
Ours        & UGVPoseNet*        & \textbf{26.65} & 89.73  & 86.75 \\  
Ours        & UGVPoseNet         & 21.89 & \textbf{93.33}  & \textbf{92.72} \\ \hline  
\end{tabular}
}
\label{tab:speed}
\end{center}
\end{table}

\subsection{Runtime Analysis}
To analyze the runtime performance of the proposed method, we evaluate the actual inference speed on a mobile device equipped with an NVIDIA GeForce GTX-1070MQ GPU. The raw image size is $1080 \times 1920$, which resized to $360 \times 640$ during inference.
We compare the inference speed of our method with other start-of-the-art methods.
We first compare our method with OpenPose~\cite{OpenPoseTPAMI}. We also replace the backbone of our framework with other networks such as Xception~\cite{chollet2017xception}, ResNet~\cite{he2016deep}, MobileNetV2~\cite{sandler2018mobilenetv2}, and ShuffleNetV2~\cite{ma2018shufflenet}.
We measure the inference time cost for each of them. The results are reported in Table~\ref{tab:speed}.
The runtime consists of two major parts, namely, CNN processing time and structure assembling time.
The latter does not influence much of the overall runtime because it is two orders of magnitude less than the CNN processing time.
In the experiments, we compare the overall inference time including both.
Our UGV-KPNet achieves a speed of 21.89FPS, and our UGV-KPNet* can reach a speed of 26.65FPS. 
In contrast, OpenPose has a much lower running speed of 2.01FPS. Our model also runs faster than the counterparts using other backbones.
In the automatic recovery process, the moving speed of the UGV is 0.5m/s.
Therefore, our system can support the UGV to obtain a position and attitude update at least every 0.025m of movement, fulling the requirement of real-time navigation.

\subsection{Evaluation of the 6-DoF pose estimation}

The 6-DoF pose of the UGV is estimated by solving the PnP problem through the keypoints predicted by the UGV-KPNet.
Then, we can get the projected keypoints in the 2D image using the estimated  6-DoF pose. 
We experiment with two methods, namely, EPnP~\cite{lepetit2009epnp} and Levenberg-Marquardt algorithm (LM)~\cite{roweis1996levenberg}. 
The performance is evaluated by measuring the average distance between the projected keypoints and the ground truth keypoints.

The results are shown in Table~\ref{tab:sys_error}, the error of EPnP is 12.89 pixels, while that of LM is only 1.98 pixels. Note that our keypionts detection error is 2.20 pixels, which is very close to the reprojection error using LM method.
Given the robust performance of the LM algorithm, we employ it in the pose estimation module of our system.
%

\begin{table}[!tb]
\caption{
Comparison results of different methods for 6-DoF pose estimation.
}
\begin{center}
\begin{tabular}{c|c|c}
\hline
Methods & EpnP~\cite{roweis1996levenberg} & LM~\cite{lepetit2009epnp} \\ 
\hline
Error (pixel)    &   12.89        &    1.98       \\
\hline
\end{tabular}

\label{tab:sys_error}
\end{center}
\end{table}


\section{Conclusion}\label{sec:Conclusion}
This paper presents an effective solution for the task of UGV autonomous recovery with a vision system. We have proposed the UGV-KPNet to detect the keypoints for UGV with the image captured by a monocular camera, which is used to estimate the reliable 6-DoF pose of the UGV in real-time. We have introduced several improvements in UGV-KPNet over existing methods, including multi-resolution subnetworks, multi-level connection, improved shuffle block, and residual adaptor. We also created a large-scale real-world dataset for UGV keypoints detection. We have comprehensively evaluated the proposed method on the large-scale real-world dataset, and the method achieves state-of-the-art results in terms of both accuracy and efficiency.
We believe the method introduced in this paper can provide a valuable reference for other robot keypoint positioning tasks in the field of robotics.


\section*{Acknowledgment}

This work was supported by the National Natural Science Foundation of China under Grants 61773210 and 61603184, and by the EPSRC Programme Grant Seebibyte EP/M013774/1.


%





\ifCLASSOPTIONcaptionsoff
  \newpage
\fi



\bibliographystyle{IEEEtran}
\bibliography{main}
%



%




\end{document}